\documentclass[runningheads]{llncs}
\usepackage{graphicx}
\usepackage{xcolor}
\usepackage{CJKutf8}
\newcommand{\chinese}[1]{\begin{CJK}{UTF8}{bsmi}#1\end{CJK}}
\usepackage{newunicodechar}
\newunicodechar{理}{\-} 
\newunicodechar{疗}{\-} 
\usepackage{comment}

\begin{document}
\title{Fine Grained Human Evaluation for English-to-Chinese Machine Translation: A Case Study on Scientific Text}
%
%
\author{Ming Liu\inst{1,2} 
He Zhang\inst{2}  
Guanhao Wu\inst{2}}
\institute{Deakin University, Melbourne, Australia \\
\email{m.liu@deakin.edu.au}\\
 \and
Zhongtukexin Co. Ltd. , Beijing, China\\
\email{zhanghe12@tsinghua.org.cn, guanhao.wu@outlook.com}}

%

%
\maketitle              
\begin{abstract}
Recent research suggests that neural machine translation (MT) in the news domain has reached human-level performance, but for other professional domains, it is far below the level. In this paper, we conduct a fine-grained systematic human evaluation for four widely used Chinese-English NMT systems on scientific abstracts which are collected from published journals and books. Our human evaluation results show that all the systems return with more than 10\% error rates on average, which requires much post editing effort for real academic use.  Furthermore, we categorize six main error types and and provide some real examples. Our findings emphasise the needs that research attention in the MT community should be shifted from short text generic translation to professional machine translation and build large scale bilingual corpus for these specific domains. 

\keywords{Machine Translation  \and Human Evaluation \and Chinese English Translation \and Metrics}
\end{abstract}
\section{Introduction}
The development of neural MT benefits from a large amount of bilingual data, English-to-Chinese MT is often considered as a high resource setting as various types of parallel sentences have been collected, such as WikiMatrxi \cite{schwenk2019wikimatrix}, the United Nations Parallel Corpus \cite{ziemski2016united}, the UM Corpus \cite{tian2014corpus}. However, evaluation for English-Chinese Machine Translation still larges relies on automatic metrics like BLEU (BiLingual Evaluation Understudy) \cite{papineni2002bleu}. Similar metrics include the NIST metric (National Institute of Standards and Technology) \cite{doddington2002automatic} and METEOR (Metric for Evaluation of Translation with Explicit Ordering) \cite{banerjee2005meteor}, which is either precision or F-score based. Other metrics are based on the error rate and the Levenshtein distance, such as the WER (Word Error Rate) score and the improved versions of this metric – i.e. PER (Position- independent Word Error Rate) \cite{popovic2007word}, TER (Translation Edit Rate or Translation Error Rate) \cite{agarwal2008meteor} and HTER (Human-targeted Translation Edit Rate) \cite{snover2006study}. More recently, embedding-based metrics show much more correlation with human judgement, such as BertScore \cite{zhang2019bertscore}, chrF \cite{popovic2015chrf}, YISI-1 \cite{lo2019yisi}, ESIM \cite{chen2016enhanced}.

In comparison, human evaluation for MT often consider the following factors: i) adequacy and fluency, ii) ranking the outputs of different systems at the system level, iii) post editing effort, i.e. how much effort does it take for a translator (or even monolingual) to ``fix" the MT output so it is ``good". iv )task-based evaluation: was the performance of the MT system sufficient to perform a task. GENIE \footnote{https://genie.apps.allenai.org} provides more accurate assessment of progress in general text generation tasks, which use human evaluation of the entries, gathered dynamically using crowdsourcing (Amazon Mechanical Turk).

In this paper, we aim to conduct an extensive human evaluation for the outputs of four English-Chinese MT systems based on scientific abstracts from biomedical and other science engineering domains. Meanwhile, we manually carry out an error analysis according to the Multidimensional Quality Metrics (MQM) framework \cite{lommel2014multidimensional}, the analysis shows the performance current English-to-Chinese MT systems for scientific domain is far below the average level compared to the news domain. The main contributions of this paper can be summarised as follows:
\begin{itemize}
    \item We develop several MQM-based metrics for English-to-Chinese MT systems, our metrics consider both sentence and document level translation quality.
    \item We conduct, to the best of out knowledge, the first fine-grained human evaluation and error analysis for English-to-Chinese MT in the scientific domain.
    \item Our evaluation results suggest that current English-Chinese MT models return more than 10\% error rate, which causes the reliability problem in the scientific domain, researchers still need much post editing effort to revise the translated text.
\end{itemize}
The rest of the paper is arranged as follows: Section 2 presents the data collection and construction process, and outlines the four English-to-Chinese MT systems that we evaluate. Section 3 presents the detailed human evaluation setup and metrics. Section 4 shows the results. Section 5 analyzes the common error types and gives a few examples. Section 6 briefly mentions the related work, Section 7 list several limitations of this work, followed by a conclusion and suggestions for future work in Section 8.
\section{Scientific Bilingual Data and English-to-Chinese Machine Translation Systems}
\subsection{Data sets}
We collect our data from three different sources:
Biomedical journals, biomedical books, full subject scientific text. The biomedical journals and books are sampled from PubMed \footnote{https://pubmed.ncbi.nlm.nih.gov}, which is a free search engine accessing primarily the MEDLINE database of references and abstracts on life sciences and biomedical topics. The full subject scientific text are abstracts sampled from arXiv \footnote{https://arxiv.org}, which provides free research papers for eight domains such as physics, mathematics, computer science, etc. Table \ref{tab:dataset} shows the basic statistics of our data set. 
\begin{table}
\caption{Data set statistics}\label{tab:dataset}
\centering
\begin{tabular}{|l|l|l|l|}
\hline
Type &   Source & Size & Description  \\
\hline
Biomedical journals & PubMed  & 10 & abstracts from PubMed journals\\
Biomedical books &  PubMed & 10 & preface from PubMed books\\
Full subject &  arXiv & 10 & abstracts from arXiv \\
\hline
\end{tabular}
\end{table}
For all the 20 biomedical text, we ask master students from medical schools to write the translated Chinese. Specifically, these students were suggested to use the ``Machine Translation + Post Editing" way to develop the translations, i.e., they can use any available MT systems but need to rewrite the MT outputs until they are satisfied with the final translations. For the rest 10 full subject abstracts, similar steps were conducted by master students from corresponding areas. We will open source the annotated translations for future research.

\subsection{Chinese-English Translation Models}
Four open source Chinese-English Machine Translation models are selected for the human evaluation, including Helsinki-mt-zh-en \footnote{https://huggingface.co/Helsinki-NLP/opus-mt-en-zh}, XiaoNiuTrans \footnote{https://niutrans.com}, BaiduTrans \footnote{https://fanyi.baidu.com}, GoogleTrans \footnote{https://pypi.org/project/google-trans-new/}. 
Helsinki-mt-en-zh is a pre-trained transformer model with PyTorch, the model has 6 encoder layers and 6 decoder layers, the maximum input lenth is 512 and the vacabulary size is 65000. More configuration about Helsinki-mt-zh-en can be found in the Huggingface model files \footnote{https://huggingface.co/Helsinki-NLP/opus-mt-zh-en/blob/main/config.json}.

The rest three MT systems are black box MT models. For XiaoNiuTrans and BaiduTrans,  we used the commercial web APIs. For GoolgeTrans, we used a Google-Trans-New python library, which is more stable than the web services, but the maximum character limit on a single text is 15k.

\section{Human Evaluation Methodology}
We define our evaluation metrics based on the MQM framework developed at the QT-LaunchPad project \cite{lommel2014multidimensional}, after reviewing a few translation quality evaluation frameworks.  Most of the focus on MQM has gone into defining analytic methods that seek to identify specific errors in a translation in order to quantify them. They thus focus on portions of the text and rely on the positive identification of specific errors. In such analytic methods the errors are seen as deviations from a text that fully complied with specifications and “errors” that do not lead to a deviation from specifications are not counted: for example, if specifications for a service manual state that stylistic concerns are not important, poor style would not be counted as an error, even though incorrect terminology would be.  

\subsection{Evaluation Metrics}
To make extensive human evaluation towards the abstract translation, we extend MQM-framework and define both sentence and document level evaluation metrics, i.e., we ask the translation model to predict the target text either at sentence or document level. The motivation is to evaluate if any of these four translation systems has integrated document level MT. We not only consider surface level metrics such as Literal Translation Rate, but also look at linguistic and semantic metrics (e.g. Coherence and Error Rate). The following lists both sentence and document level human evaluation metrics. 
\subsubsection{Sentence level metrics}
\paragraph{Sentence Error Rate (SER):} the percentage of wrongly translated sentences. The sentences are translated one by one without knowing any context information. Section 6 will list common errors discovered during our evaluation.
\paragraph{Fluency:} The fluency level within a single sentence. The rating is ranged from 1 to 5, with 1 as the least satisfied and 5 as the most satisfied.
\paragraph{Sentence Coherence (SC):} The coherence between the sentences. We use this metric not only to see if the translation is good or not, but also check if there will be one-to-many or many-to-one translation patterns in scientific text. The rating is ranged from 1 to 5, with 1 as the least satisfied and 5 as the most satisfied.
\paragraph{Literal Translation Rate (LTR):} The proportion of sentences which use direct word by word translation. We find that literal translation is more efficient for scientific text as the senses for professional words are quite limited compared to other domains like business and poem. For example, given the English sentence ``The new crown epidemic is so fierce". The translation `` \chinese{疫情是如此猛烈}" is regarded as more literal than the text ``\chinese{疫情猛如龙虎}".
\paragraph{Sentence Over Translation Rate (SOT):} The proportion of sentences which are over translated. There are two scenarios for over translation: one is the MT model repeatedly translates a word or phrase, the other is the model create some new context which never appeared in the source language. 
\paragraph{Sentence Score:} An average overall sentence translation quality score consider all the above sentence metrics, with 0 as strongly disagree and 1 as full score.

\subsubsection{Document level metrics}
\paragraph{Document Error Rate (DER):} the percentage of wrongly translated sentences. 
\paragraph{Coherence:} The coherence level for a given translated document, this metric is more focused on the context level. The rating is from 1 to 5, with 1 as least satisfied and 5 as most satisfied. Annotators are required to give the score after they finishing reading the whole translated abstract. 
\paragraph{Generality:} If the main meaning of the source abstract is translated into the target side. The rating is from 1 to 5, with 1 as least satisfied and 5 as most satisfied. In the scenario of scientific text, annotators were asked to check if the translation covers the main contribution of the original paper.
\paragraph{Logic:} Whether the structure of the source abstract is kept. The rating is from 1 to 5, with 1 as least satisfied and 5 as most satisfied. It is found that most abstract from the biomedical and science engineering domains exhibit the background-methodology-result pattern.  
\paragraph{Missing Rate:} The proportion of sentences which are not translated within the whole source text. 
\paragraph{Document Over Translation Rate:} The number of sentences which are over translated divided by the total number of sentences in the source document. It is noticed that there are some urls or email adresses which are not necessarily to be translated, we consider these ones as over translation. Different from sentence over translation which considers the semantic level, document over translation cares more about the structure level. 
\paragraph{Sentence Segmentation Correctness Rate:} The correctness level for complex sentence translation, in which 
\paragraph{Context Score:} An average overall document translation quality score consider all the above document metrics, with 0 as strongly disagree and 1 as full score.

\subsection{Evaluation Setup}
We use translate5 \footnote{http://www.translate5.net}, an open-source web-based tool, as the annotation environment. translate5 was installed on a cloud server, so that it could be accessed remotely by annotators. The source text and reference translation are provided next to the four NMT translations. For each piece of translation outputs, we ask two annotators to read all the four systems' translations and discuss with each other. After the discussion, they would together assign scores according to the evaluation metrics. In case the two annotations disagree with each other in some translations, an average score from them will be used as the final evaluation. 

\section{Results}
Table \ref{tab:result_sentence} and Table \ref{tab:result_doc} shows the results from sentence and document level human evaluation respectively. In Table \ref{tab:result_sentence}, for the sentence level translations, it can be seen that the pre-trained Helsinki-en-zh model returns much higher error rate than other three commercial systems on all the datasets. Helsinki-en-zh also shows more than 50\% literal translation rate and less fluency ratings. Specifically, XiaoNiuTrans and BaiduTrans performs on par with each other in the biomedical translations, while GoogleTrans performs better for arXiv abstract translation. The sentence coherence from all four systems' translation is similar to each other. In Table \ref{tab:result_doc}, for the document level translations, we can see the BLEU scores returned by these four systems are around 0.1, which is much lower than that of the news domain (typically from 0.3 to 0.4). Helsinki-en-zh again exhibits more errors: 43.33\% for biomedical journals, 66.43\% for biomedical books and 13.99\% for arXiv abstracts. XiaoNiuTrans and BaiduTrans show lower error rate than GoogleTrans at the document level. For other metrics such as coherence, abstractiveness, logics and missing rate, XiaoNiuTrans and BaiduTrans perform better than Helsinki-en-zh and GoogleTrans.

\begin{table}
\caption{Sentence Level Human Evaluation Results}
    \centering
    \label{tab:result_sentence}
    \begin{tabular}{|c|c|c|c|c|c|c|c|}
    \hline
         Source & Model & ER & Fluency & SC& LTR & OTR & Sentence Score \\
    \hline
         & Helsinki & 36.67\% & 3.67 & 5 & 66.67\% & 0 & 0.74 \\
         Bio  & XiaoNiuTrans & 10.00\%	& 5	& 5	& 46.67\% &	0 & 0.89 \\ 
         journals & BaiduTrans & 10.00\% &	4.33 &	5 &	73.33\%	& 0 & 0.81 \\
         & GoogleTrans & 13.33\% &	4.33 &	5 &	60.00\%	& 6.67\% &	0.81 \\
    \hline 
         & Helsinki & 58.30\%	& 2.75 &	5 &	51.43\%	& 7.14\% &	0.68 \\
         Bio  & XiaoNiuTrans & 13.21\% &	4.5 &	5 &	48.75\%	 &0 &	0.86 \\
         books & BaiduTrans & 10.71\%	& 5	 & 5 &	52.50\% &	0 &	0.87 \\
         & GoogleTrans & 13.33\% &	4.33 &	5 &	60.00\%	& 6.67\% &	0.81 \\
    \hline 
         & Helsinki & 4.90\% &	5 &	5 &	100.00\% &	0 &	0.79 \\
         arXiv & XiaoNiuTrans & 11.11\%	& 5	& 5 & 100.00\%	& 0 &	0.78 \\
         abstracts & BaiduTrans & 11.11\%	& 5 & 5 &	100.00\% &	0	& 0.78 \\
         & GoogleTrans & 0.00\%	 & 5.00 & 	5 &	0 &	0 &	1.00 \\
    \hline 
    \end{tabular}
\end{table}
\begin{table}[]
\caption{Document Level Human Evaluation Results}
\label{tab:result_doc}
    \centering
    \begin{tabular}{|c|c|c|c|c|c|c|c|c|c|c|}
    \hline
         Source & Model & BLEU & DER & DC & DA & Logic & MR & OTR & SSC & Context \\
         \hline 
         & Helsinki &0.109 &43.33\% &3.33 &4 &4 &16.67\% &3.33\% &4 &0.78\\
         Bio& XiaoNiuTrans & 0.101 & 10\% &4.33 & 5 &5 &6.67\% &0 &5 & 0.96\\
         journals& BaiduTrans & 0.108 &10.00\% &4.00 &5 &4.67 &6.67\% &0 &5 & 0.94\\
         &GoogleTrans &0.128 &13.33\% &4.33 &4.67 &4.67 &6.67\% &6.67\% &4.67 &0.91\\
         \hline
         &Helsinki &0.100 &66.43\% &1.5 &1.5 &1.25 &36.43\% &11.07\% &1.75 &0.44\\
         Bio&XiaoNiuTrans  &0.114 &13.21\% &4.5 &4.75 &4.5 & 0 & 0 &4.76 & 0.94\\
         books&BaiduTrans &0.117 &14.29\% &5 &4.75 &4.75 &0 &0 &5 &0.97\\
         &GoogleTrans &0.126 &29.29\% &4.75 &4.50 &4.75 &0 &0 &4.50 & 0.92\\
         \hline
         &Helsinki &0.101 &13.99\% &4.67 &5 &5 &3\% &0 &5 &0.97\\
         arXiv&XiaoNiuTrans &0.112 & 11.11\% & 4.67 & 5 & 5 & 0 & 0 &5 &0.97\\
         abstracts&BaiduTrans &0.113 &11.11\% &5 &5 &5 &0 &0 &5 &0.98\\
         &GoogleTrans &0.125 &0 &5 &5 &5 &0 &0 &5 &1\\
         \hline
    \end{tabular}
\end{table}
In general, we find that Helsinki-en-zh is not a reliable translation model without fine tuning on the target domain. XiaoNiuTrans and BaiduTrans does better than GoogleTrans in the biomedical domain while GoogleTrans shows good performance for arxiv abstract translation. Still, all these four systems' performance are far from professional requirement in the real academic world.

\section{Common Error Analysis}
We conduct further error analysis for the translated text, and categorize the main errors into six different types:
\paragraph{No Translation Error: }  Helsinki-en-zh suffered more from this problem than other systems. For example, `` Chagas disease, caused by Trypanosoma cruzi" was translated as ``Chagas \chinese{疾病是由} Trepanosoma cruzi \chinese{引起的}". We are aware that for some biomedical abbreviations or professional jargon there is no need to do the translation (e.g. CT image). However, for the disease name in the above example, we anticipate the main reason is there is no such domain specific words in the vocabulary, since the vocabulary size of Helsinki-en-zh is 65k. There are also cases that some clauses or sub-sentences which are not translated. 
\paragraph{Mistranslation Error:} Mistranslation is another major issue for professional word translation. For biomedical text, most mistranslation happend for medical entities, for example, ``For either therapeutic goal, understanding the mechanisms of cell death  becomes paramount." was translated as ``\begin{CJK*}{UTF8}{gbsn}对任何理疗来说了解细胞死亡机制都很重要\end{CJK*}". The word ``therapeutic" should be translated as ``\begin{CJK*}{UTF8}{gbsn}治疗\end{CJK*}" rather than 理疗\begin{CJK*}{UTF8}{gbsn}
``理疗" 
\end{CJK*}. 
\paragraph{Addition Error:} 
We find Helsinki-en-zh and GoogleTrans returned some addition errors, i.e., they over translated some words by adding some unnecessary translations.
For example, ``Insulin resistance, underlying factors in obesity-related diseases" was translated as ``  \begin{CJK*}{UTF8}{gbsn}胰岛素抗药性炎症、肥胖相关疾病的基本因素\end{CJK*}". In the Chinese translation, the word ``  \begin{CJK*}{UTF8}{gbsn}炎症\end{CJK*}" is redundant and useless. This is a rare problem but still causes some fact issues for the translations. 
\paragraph{Repetition Error: } 
Helsinki-en-zh also returned many repetitive errors. 
For example, ``Among the molecular targets, enzymes of the sterol pathway.." was translated as ``\begin{CJK*}{UTF8}{gbsn}在分子的分子目标目标目标中...\end{CJK*}". The main reason is Helsinki-en-zh is a small sized model, there exist too many words predicting the same word as the subsequent word with high probability in the training time. Consequently, it is easy to go back to that world and form repetitions. A theoretical analysis of the repetition problem in text generation was mentioned in \cite{fu2020theoretical}.
\paragraph{Grammar Error:} Even though English and Chinese have the same ``SVO" pattern, we still find some grammar errors. These errors do not have a big impact on the semantics but return unnatural translations.
\paragraph{Punctuation Error:} The use of comma is a main punctuation error. Helsinki-en-zh and GoogleTrans return such errors when commas are inserted into the translated sentence. 
For the above six types of errors, the first three types cause incorrect meaning understanding for the source language, and the other three errors bring linguistic and structure difficulty for the target langauge.

\section{Related Work}
Neural MT generally follows an encoder decoder architecture, in which the encoder builds a continuous representation  for  the  source  sentence  while  the decoder  is  a  neural  language  model  conditioned on the encoder output. The parameters are learned jointly  to  maximize  the  likelihood  of  the  target sentences given the corresponding sentences with a parallel corpus. At prediction time, either greedy or beam search is used to generate the target sentence from left to right. Various architectures have been proposed to improve  the quality  of  neural  machine  translation.This involves recurrent networks \cite{bahdanau2014neural}, convolutional networks \cite{gehring2016convolutional} and transformer networks \cite{vaswani2017attention}.   Attention has shown great help for these neural architectures,  which  includes  self-attention \cite{tang2018self},  multi-hop attention \cite{iida2019attention} and multi-head attention \cite{alkhouli2018alignment}. 

Automatic evaluation for Neural MT can be categorized into the following five approaches: precision-based, such as BLEU \cite{papineni2002bleu}, NIST \cite{przybocki2009nist}.  F-score-based: such as Meteor \cite{lavie2009meteor}.  Error rates, such as WER \cite{he2011word}, TER \cite{agarwal2008meteor}, PER \cite{popovic2007word}. Using syntax or semantics, such as PosBleu \cite{popovic2009syntax}, Meant \cite{lo2011meant}, DepRef \cite{graham2015accurate}. 
Embedding based, such as BertScore \cite{zhang2019bertscore}, chrF \cite{popovic2015chrf}, YISI-1 \cite{lo2019yisi}, ESIM \cite{chen2016enhanced}. 
Despite the popularity of such automatic metrics, human evaluation is valuable when it comes to informing experts to further improve MT models. \cite{escribe2019human} provided a framework for error categorisation and conducted a comparative analysis of the raw translation output and the post-edited version with the purpose of identifying recurring patterns of errors. \cite{ye2020fine} presented a fine-grained human evaluation to compare the Transformer and recurrent approaches to neural MT, in which they conducted an error annotation using a customised error taxonomy on the output of state-of-the- art recurrent and Transformer-based MT systems on a subset of WMT2019’s news test set. Their annotation shows the best Transformer system results in a 31\% reduction of the total number of errors compared to the best recurrent systems. \cite{popovic2020informative} proposed a method for manual evaluation of MT output based on marking actual issues in the translated text, in which  evaluators are not assigning any scores, nor classifying errors, but marking all problematic parts (words, phrases, sentences) of the translation. 

In the recent WMT2020 Metrics Shared Task \cite{mathur2020results}, participants were asked to score the outputs of the translation systems competing in the WMT20 News Translation Task with automatic metrics. Ten research groups submitted 27 metrics, four of which are reference-less ``metrics”. The organizer computed five baseline metrics, including SENTBLEU, BLEU, TER and CHRF using the SacreBLEU scorer. All metrics were evaluated on how well they correlate at the system, document and segment level with the WMT20 official human scores.
\section{Limitations}
Our work has a few limitations, we list them as follows:
\paragraph{Size of corpus: } The first limitation of this work corresponds to the size of the corpus analysed. Even though we randomly sampled the 300 abstracts from biomedical, science and engineering domains, verifying whether the results obtained in this study apply to the whole domain constitute a valuable analysis. 
\paragraph{Language pair:} We focused on English to Chinese translation which is unidirectional, the other direction can also be evaluated. Meanwhile, the work can be extended to low-resource language pairs such as Mongolian and Tibetan.
\paragraph{MT models:} We only evaluated one open source and three commercial English-Chinese MT systems, which may not the state of the art models. In the recent WMT2020 competition\footnote{http://www.statmt.org/wmt20/}, more advanced MT models have been developed in the track of news and biomedical translation. It is noticed that HelsinkiNLP and XiaoNiuTrans also participated the WMT2020 competition. 

\paragraph{Text genre} The source text we used is limited to the scientific domains. We only focus on the biomedical and science engineering areas, since these types of text are more objective. Areas from art and poems could also be considered, as there will be more emotions and ambiguities from the source side. 

\paragraph{Annotator} This work is based on human evaluation only, with discussions between two annotators in the corresponding field. A suggestion for future research is to involve more emulators and conduct quality control. Especially, inter-rate agreement testing is important when assessing the final results. It would also appear relevant to analyse the same corpus with automatic metrics and to compare the results thus obtained with the findings of the present study.

\paragraph{Translation error types} The translation error types adopted for this research also constitutes a limitation. Since in the research community people care more about the fact of the source context, more emphasize could be give to other linguistic factors such as word order, the translation of functional words, etc. The error types can be further detailed based on the MQM-complicant error hierarchy as shown in \cite{escribe2019human,ye2020fine}.

\section{Conclusion and Future Work}
This paper presents the human evaluation results for the outputs of four English-Chinese MT systems based on scientific abstracts from biomedical and other science engineering domains. We design sentence and document level metrics to evaluate the translation outputs, meanwhile, we manually carry out an error analysis and summarizes six types of main errors, the analysis shows the performance current English-to-Chinese MT systems for scientific domain is far below the average level compared to the news domain.  

In the future work, apart from the directions which solve the above limitations, we plan to do build a large scale English Chinese bilingual corpus for scientific text. Meanwhile, we will train the state of art transformer models based on this corpus and evaluate the model performance accordingly.
%
%
%
\bibliographystyle{splncs04}
\bibliography{mybibliography}

\end{document}